%% file: paper.tex
\documentclass[review]{elsarticle}

\usepackage{lineno,hyperref}
\usepackage{color, colortbl, amsmath}
\usepackage[linesnumbered,ruled]{algorithm2e}
\usepackage{xcolor,hhline}
\usepackage{multirow}

\modulolinenumbers[5]

\journal{Journal of \LaTeX\ Templates}

\begin{document}

\begin{frontmatter}

\title{A Metaheuristic-Driven Approach to Fine-Tune Deep Boltzmann Machines}

\author[1]{Leandro Aparecido Passos}	
\author[2]{Jo\~{a}o Paulo Papa}

\address[1]{Department of Computing, Federal University of S\~ao Carlos\\
	Rod. Washington Luís, Km 235, S\~ao Carlos, 13565-905, Brazil\\
	leandro.passosjr@dc.ufscar.br}

\address[2]{Department of Computing, S\~ao Paulo State University\\Av. Eng. Luiz Edmundo Carrijo Coube, 14-01, Bauru, 17033-360, Brazil\\joao.papa@unesp.br}

\begin{abstract}
Deep learning techniques, such as Deep Boltzmann Machines (DBMs), have received considerable attention over the past years due to the outstanding results concerning a variable range of domains. One of the main shortcomings of these techniques involves the choice of their hyperparameters, since they have a significant impact on the final results. This work addresses the issue of fine-tuning hyperparameters of Deep Boltzmann Machines using metaheuristic optimization techniques with different backgrounds, such as swarm intelligence, memory- and evolutionary-based approaches. Experiments conducted in three public datasets for binary image reconstruction showed that metaheuristic techniques can obtain reasonable results.
\end{abstract}


\end{frontmatter}


\input{sections/introduction.tex}

\input{sections/theoretical.tex}

\input{sections/optimization.tex}

\input{sections/methodology.tex}
\input{sections/experiments.tex}

\input{sections/conclusion.tex}

\section*{Acknowledgments}
\begin{sloppypar}
The authors are grateful to FAPESP grants \#2013/07375-0, \#2014/12236-1, and \#2016/19403-6, and CNPq grants \#306166/2014-3, and \#307066/2017-7. This material is based upon work supported in part by funds provided by Intel\textsuperscript{\textregistered} AI Academy program under Fundunesp Grant No.2597.2017. This study was financed in part by the Coordena\c{c}\~ao de Aperfei\c{c}oamento de Pessoal de N\'ivel Superior - Brasil (CAPES) - Finance Code 001
\end{sloppypar}

\bibliography{paper}

\end{document}

%% file: sections/introduction.tex
\section{Introduction}
\label{s.intro}

Restricted Boltzmann Machines (RBMs)~\cite{Hinton:02, passosTESE:2018} are probabilistic models that employ a layer of hidden binary units, also known as latent units, to model the distribution of the input data (visible layer). Such models have been applied to deal with problems involving images~\cite{larochelle2007empirical}, text~\cite{salakhutdinov2009semantic}, detection of malicious content~\cite{fiore2013network,SilvaIJCSIS:16}, and several diseases diagnosis~\cite{pereiraCAIP:2017,khojastehCBM:2019,passosJVCIR:2019}, just to cite a few. Moreover, RBMs are also used for building deep learning architectures, such as Deep Belief Networks (DBNs)~\cite{hinton2006fast} and Deep Boltzmann Machine (DBM)~\cite{salakhutdinov2009deep,passosNPL:2018}, where the main difference is related to the interaction among layers of RBMs.  

Deep Learning techniques have been extensively used to deal with tasks related to signal processing and computer vision, such as feature selection~\cite{ruangkanokmas2016deep}~\cite{SohnNIPS:15}, face~\cite{taigman2014deepface}~\cite{DuongCVPR:15} and image reconstruction~\cite{dong2014learning}, multimodal learning~\cite{srivastava2012multimodal}, and topic modeling~\cite{hinton2009replicated}, among others. Despite the outstanding results obtained by these models, an intrinsic constraint associated with deep architectures is related to their complexity, which can become an insurmountable problem due to the high number of hyperparameters one must deal with. The present work focuses on this problem.

Some works have recently modeled the issue of hyperparameter fine-tuning as a metaheuristic optimization task. Such techniques show up as an interesting alternative for such a task since they do not require computing derivatives of hundreds of parameters as usually happen with standard optimization techniques, which is not recommended for high-dimensional spaces. Papa et. al.~\cite{PapaGECCO:15}, Rosa et al.~\cite{rosa2016learning}, and Passos et al.~\cite{passosiRBM:2017} are among the first to introduce metaheuristic-driven optimization in the context of RBMs, DBNs, and Infinity Restricted Boltzmann Machines (iRBMs) hyperparameter fine-tuning, obtaining more precise results than the ones achieved using some well-known optimization libraries in the literature~\footnote{Notice the context of hyperparameter fine-tuning stands for a proper selection of the network's input values, such as the number of hidden units and the learning rate, among others, rather than optimizing the bias and weights of the model.}. 

Recently, Passos et al.~\cite{passosSACI:2018} proposed to employ metaheuristic approaches in the context of DBM hyperparameter optimization. However, the work deals only with Harmony Search~\cite{Geem:09} and Particle Swarm Optimization~\cite{Kennedy:01} techniques. Moreover, the paper presents a shallow discussion regarding the experimental results. Therefore, in this work, we considered DBM hyperparameter fine-tuning in the context of music-inspired, swarm-based and differential evolution algorithms, employing seven different techniques: Improved Harmonic-Search (IHS)~\cite{mahdavi2007improved}, Adaptive Inertia Weight Particle Swarm Optimization (AIWPSO)~\cite{yu2009adaptive}, Cuckoo Search (CS)~\cite{yang2009cuckoo}, Firefly Algorithm (FA)~\cite{Yang:2010ffa}, Backtracking Search Optimization Algorithm (BSA)~\cite{civicioglu:13}, Adaptive Differential Evolution (JADE)~\cite{zhang:09}, and the Differential Evolution Based on Covariance Matrix Learning and Bimodal Distribution Parameter Setting Algorithm (CoBiDE)~\cite{wang:14}. Furthermore, all techniques are compared with a random search for experimental purposes. Additionally, this work provides a more detailed experimental section, considering a statistical similarity and time consumption comparison. Finally, the application addressed in this paper concerns the task of binary image reconstruction, and for that purpose, we considered three public datasets.

In a nutshell, the main contribution of this paper is to introduce a detailed analysis considering metaheuristic optimization to the context of DBM hyperparameter fine-tuning, as well as to foster the research towards such area. Additionally, we provided an extensive experimental evaluation with distinct learning algorithms over a different number of layers. As far as we are concerned, we have not observed any study with such level of details. The remainder of this paper is presented as follows. Section~\ref{s.theoretical} presents the theoretical background related to RBMs, DBNs, and DBMs. Section~\ref{s.fineTuning} introduces the main foundations related to the metaheuristic optimization techniques employed in this work. Sections~\ref{s.methodology} and~\ref{s.experiments} present the methodology and experiments, respectively, and Section~\ref{s.conclusions} states conclusions and future works.

%% file: sections/theoretical.tex
\section{Theoretical Background}
\label{s.theoretical}

\subsection{Restricted Boltzmann Machines}
\label{ss.rbm}

\begin{figure}[!ht]
\centerline{\begin{tabular}{c}
\includegraphics[width=8.0cm ]{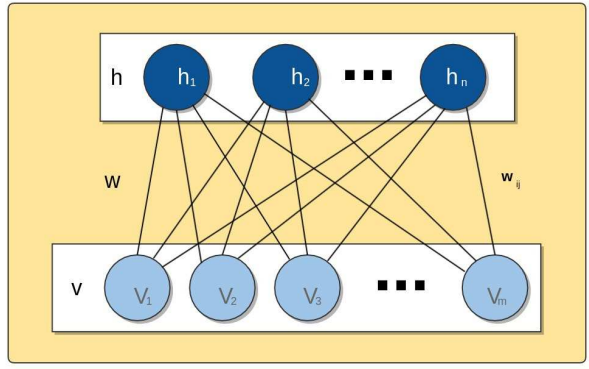} \\
\end{tabular}}
\caption{The RBM architecture.}
\label{f.rbm}
\end{figure}

Restricted Boltzmann Machines are stochastic models composed a visible and a hidden layer of neurons, whose learning procedure is based on the minimization of an energy function. A vanilla  architecture of a Restricted Boltzmann Machine is depicted in Figure~\ref{f.rbm}, which comprises a visible layer $\textbf{v}$ with $m$ units and a hidden layer $\textbf{h}$ with $n$ units. Furthermore, $\textbf{W}_{m\times n}$ stand for a real-valued matrix that models the weights between both layers, as well as $w_{ij}$ stands for the weight between the visible unit $v_i$ and the hidden unit $h_j$.

Assuming both $\textbf{v}$ and $\textbf{h}$ as binary-valued units, i.e., $\textbf{v}\in\{0,1\}^m$ e $\textbf{h}\in\{0,1\}^n$, the energy function of models is given by:

\begin{equation}
\label{e.energy_bbrbm}
E(\textbf{v},\textbf{h})=-\sum_{i=1}^ma_iv_i-\sum_{j=1}^nb_jh_j-\sum_{i=1}^m\sum_{j=1}^nv_ih_jw_{ij},
\end{equation}
where $\textbf{a}$ e $\textbf{b}$ stand for the biases of visible and hidden units, respectively.

Since the RBM is a bipartite graph, the activations of both visible and hidden units are mutually independent, thus leading to the following conditional probabilities:

\begin{equation}
	P(\textbf{v}|\textbf{h})=\prod_{i=1}^mP(v_i|\textbf{h}),
\end{equation}
and

\begin{equation}
	P(\textbf{h}|\textbf{v})=\prod_{j=1}^nP(h_j|\textbf{v}),
\end{equation}
where

\begin{equation}
\label{e.probv}
P(v_i=1|\textbf{h})=\phi\left(\sum_{j=1}^nw_{ij}h_j+a_i\right),
\end{equation}

and

\begin{equation}
\label{e.probh}
P(h_j=1|\textbf{v})=\phi\left(\sum_{i=1}^mw_{ij}v_i+b_j\right).
\end{equation}
Where $\phi(\cdot)$ represents the sigmoid function.

Suppose the set of RBM parameters $\theta=(W,a,b)$  can be learned through a training algorithm that aims at maximizing the product of probabilities given all the available training data ${\cal V}$, as follows:

\begin{equation}
	\arg\max_{\Theta}\prod_{\textbf{v}\in{\cal V}}P(\textbf{v}).
\end{equation}
The aforementioned equation can be solved using the following derivatives over the matrix of weights \textbf{W}, and biases $\textbf{a}$ and $\textbf{b}$ at iteration $t$ as follows:

\begin{equation}
\label{e.updateW2}
\textbf{W}^{t+1}=\textbf{W}^t+\underbrace{\eta(P(\textbf{h}|\textbf{v})\textbf{v}^T-P(\tilde{\textbf{h}}|\tilde{\textbf{v}})\tilde{\textbf{v}}^T)+\Phi}_{=\Delta\textbf{W}^t},
\end{equation}

\begin{equation}
\label{e.updatea2}
\textbf{a}^{t+1}=\textbf{a}^t+\underbrace{\eta(\textbf{v}-\tilde{\textbf{v}})+\alpha\Delta \textbf{a}^{t-1}}_{=\Delta\textbf{a}^t}
\end{equation}
and

\begin{equation}
\label{e.updateb2}
\textbf{b}^{t+1}=\textbf{b}^t+\underbrace{\eta(P(\textbf{h}|\textbf{v})-P(\tilde{\textbf{h}}|\tilde{\textbf{v}}))+\alpha\Delta \textbf{b}^{t-1}}_{=\Delta\textbf{b}^t},
\end{equation}
where $\alpha$ denotes the momentum and $\eta$ stands for the learning rate. To obtain the terms $P(\tilde{\textbf{h}}|\tilde{\textbf{v}})$ and $\tilde{\textbf{v}}$, one can perform the Contrastive Divergence~\cite{Hinton:02} technique, which basically ends up performing Gibbs sampling using the training data as the visible units. In short, Equations~\ref{e.updateW2},~\ref{e.updatea2} and~\ref{e.updateb2} employ the well-known Gradient Descent as the optimization algorithm. The additional term $\Phi$ in Equation~\ref{e.updateW2} is used to control the values of matrix $\textbf{W}$ during the convergence process, and it is formulated as follows:

\begin{equation}
\label{e.theta}
	\Phi = -\lambda\textbf{W}^t+\alpha\Delta\textbf{W}^{t-1},
\end{equation}
where $\lambda$ stands for the weight decay.

\subsection{Deep Belief Networks}
\label{ss.DBN}

In a nutshell, DBNs are deep archtectures composed of a set of stacked RBMs, whose are trained in a greedy fashion using the learning algorithm presented in Section~\ref{ss.rbm}, i.e., CD and PCD. In other words, an RBM does not consider the other layers' units states while training the model at a certain layer, except that the hidden units at layer $i$ become the input units to the layer $i+1$. Suppose we have a DBN composed of $L$ layers, being $\textbf{W}^i$ the weight matrix of RBM at layer $i$. 

Hinton~\cite{HintonNC:06} proposed to consider a fine-tuning as the final step for training a DBN, aiming to adjust the matrices $\textbf{W}^i$, $i=1,2,\ldots,L$. The procedure is performed using Backpropagation or the Gradient descent algorithm. The idea is to minimize some error measure considering the output of an additional layer placed at the top of the DBN after the training procedure. The aforementioned layer is often composed of logistic units, a softmax, or even some supervised technique.

\subsection{Deep Boltzmann Machines}
\label{ss.dbm}

\begin{figure}[!ht]
\centerline{\begin{tabular}{c}
\includegraphics[width=8.0cm]{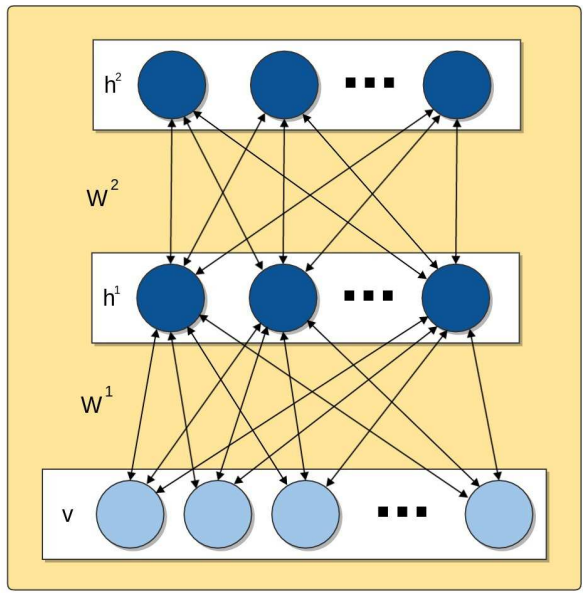} \\
\end{tabular}}
\caption{The DBM architecture with two hidden layers. Notice the double-sided arrows stand for the top-down feedback in addition to the usual bottom-up pass employed in the DBM approximate inference procedure.}
\label{f.dbm}
\end{figure}

As well as DBNs, DBMs are deep architectures able to learn more complex and intrinsic representations of the input data employing stacked RBMs. Figure~\ref{f.dbm} depicts the architecture of a standard DBM, which formulation has mild differences from the DBN~\footnote{The main difference stands in the top-down feedback used to approximate the inference procedure. Moreover, the DBM has entirely undirected connections, while the DBN has undirected connections in the top two layers only, as well as directed connections at the lower layers.}. 

The energy of a DBM  with two layers, where $\textbf{h}^1$ stands for the hidden units and $\textbf{h}^2$ stands for the visible ones in the first and second layers, can be computed as follows:

\begin{equation}
	\label{e.dbm_energy}
	E(\textbf{v},\textbf{h}^1,\textbf{h}^2)=-\sum_{i=1}^{m^1}\sum_{j=1}^{n^1}v_ih^1_jw^1_{ij}-\sum_{i=1}^{m^2}\sum_{j=1}^{n^2}h^1_ih^2_jw^2_{ij},
\end{equation}
where $m^1$ and $m^2$ stand for the number of visible units in the first and second layers, respectively, and $n^1$ and $n^2$ stand for the number of hidden units in the first and second layers, respectively. Furthermore, the weight matrices $\textbf{W}^1_{m^1\times n^1}$ and $\textbf{W}^2_{m^2\times n^2}$ encodes the weights of the connections between vectors $\textbf{v}$ and $\textbf{h}^1$, and vectors $\textbf{h}^1$ and $\textbf{h}^2$, respectively. The bias terms are dropped out for simplification purposes.

Due to its complexity, calculating the derivatives of RBM-based models becomes a prohibitive task. To deal with such a constraint, one can employ the Contrastive Divergence algorithm and sample an estimated state of the visible and hidden units. Thus, the conditional probabilities over the visible and the two hidden units are given as follows:

\begin{equation}
\label{e.probv_dbm}
	P(v_i=1|\textbf{h}^1)=\phi\left(\sum_{j=1}^{n^1}w^1_{ij}h^1_j\right),
\end{equation}

\begin{equation}
\label{e.probh2_dbm}
	P(h^2_z=1|\textbf{h}^1)=\phi\left(\sum_{i=1}^{m^2}w^2_{iz}h^1_i\right),
\end{equation}
and 

\begin{equation}
\label{e.probh1_dbm}
	P(h^1_j=1|\textbf{v},\textbf{h}^2)=\phi\left(\sum_{i=1}^{m^1}w^1_{ij}v_i+\sum_{z=1}^{n^2}w^2_{jz}h^2_z\right),
\end{equation}

Finally, the generative model can be written as follows:

\begin{equation}
	P(\textbf{v})=\sum_{\textbf{h}^1}P(\textbf{h}^1)P(\textbf{v}|\textbf{h}^1),
\end{equation}
where $P(\textbf{h}^1)=\sum_{\textbf{v}}P(\textbf{h}^1,\textbf{v})$. Further, we shall proceed with the learning process of the second RBM, which then replaces $P(\textbf{h}^1)$ by $P(\textbf{h}^1)=\sum_{\textbf{h}^2}P(\textbf{h}^1,\textbf{h}^2)$. Roughly speaking, using such procedure, the conditional probabilities given by Equations~\ref{e.probv_dbm}-\ref{e.probh1_dbm}, and Contrastive Divergence, one can learn DBM parameters one layer at a time~\cite{Salakhutdinov:12}. Later, one can apply mean-field-based learning to obtain a more accurate model.

%% file: sections/optimization.tex
\section{DBM Fine-Tuning as an Optimization Problem}
\label{s.fineTuning}

In general, Restricted Boltzmann Machines demands a proper selection of four main parameters: number of hidden units $n$, the learning rate $\eta$, the weight decay $\lambda$, and the momentum $\varphi$. Since Deep Boltzmann Machines stack RBMs on top of each other, if one has $L$ layers, then each optimization encodes $4L$ variables to be optimized. However, as the training procedure of DBMs are greedy-wise (we are not considering mean-field-based learning in this work), which means each layer is trained independently, only $4$ variables are optimized per layer.

\begin{figure}[!ht]
\centerline{\begin{tabular}{c}
\includegraphics[width=10.0cm]{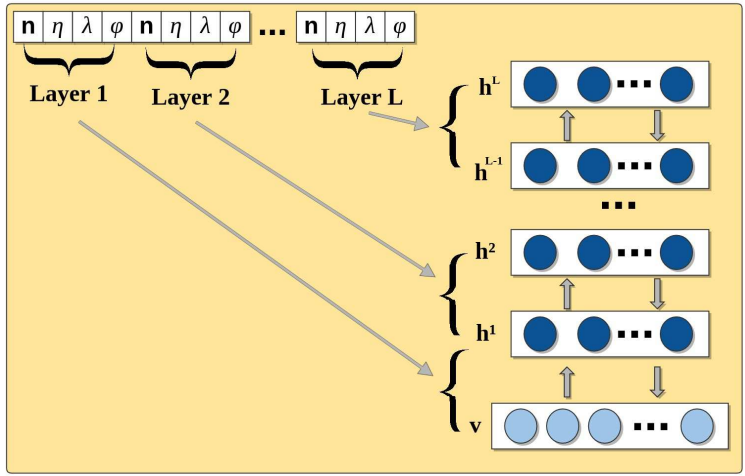} \\
\end{tabular}}
\caption{\label{f.proposed}Proposed approach to encode the decision variables of each optimization agent.}
\end{figure}

In short, the idea is to initialize all optimization techniques at random, and them the algorithm takes place. The following ranges were considered in this work parameters\footnote{The ranges used for each parameter were empirically selected based on values commonly adopted in the literature~\cite{papaQUATERNION:17, rosa2016learning, PapaGECCO:15, RodriguesBook:16, passosiRBM:2017}}: $\eta\in[0.1,0.9]$, $n\in[5,100]$, $\varphi\in[0.00001,0.01]$ and $\lambda\in[0.1,0.9]$. Aiming to fulfill the requirements of any optimization technique, one shall design a fitness function to guide the search into the best solutions. For such purpose, the mean squared error (MSE) over the training set was considered for the task of binary image reconstruction as the fitness function. Therefore, we adopted the very same methodology used by~\cite{PapaASC:15} to allow a fair comparison against the works. Figure~\ref{f.proposed} depicts the optimization model employed in this paper. In short, the approach proposed in this paper models the whole set of $4L$ decision variables as being an optimization agent. 

\begin{figure}[!htb]
\centerline{\begin{tabular}{c}
\includegraphics[width=10.3cm]{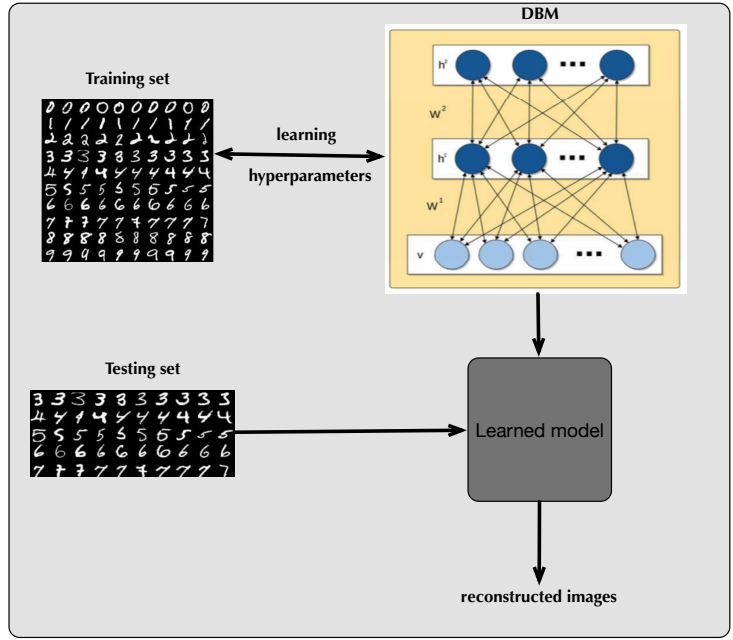} \\
\end{tabular}}
\caption{\label{f.pipeline}Proposed approach to encode the decision variables of each optimization agent.}
\end{figure}

Figure~\ref{f.pipeline} presents an overall idea of the pipeline used in this work to perform DBM hyperparameter fine-tuning. Roughly speaking, the optimization technique selects the set of hyperparameters that minimize the MSE over the training set considering a dataset of binary images as an input to the model. After learning the hyperparameters, one can proceed to the reconstruction step concerning the testing images, whose MSE is the one used to finally evaluate the metaheuristic techniques considered in this work.

\subsection{Optimization Techniques}
\label{ss.optimizationTechniques}

Below, we present a brief description of the metaheuristic techniques employed in this paper:

\begin{itemize}

\item IHS: a variant of the HS, which models the problem of function minimization based on way musicians create their songs with optimal harmonies. This approach uses dynamic values for both the Harmony Memory Considering Rate (HMRC), which is responsible for creating new solutions based on previous experience of the music player, and the Pitch Adjusting Rate (PAR), which is in charge of applying some disruption to the solution created with HMRC in order to avoid the pitfalls of local optima. Both parameters are updated at each iteration with the new values within the range $[$HMCR$_{min},$HMCR$_{max}]$ e $[$PAR$_{min},$PAR$_{max}]$, respectively. Concerning PAR calculation, the bandwidth variable \emph{(bandwidth)} $\varrho$ is used, and its values must be between $[\varrho_{min},\varrho_{max}]$.

\item AIWPSO: a variant of the PSO, which considers any possible solution as a particle (agent) in a swarm. Each agent has a position and velocity vector in the search space, as well as two acceleration constants $c_1$ and $c_2$. A fitness value is associated with each position, and after some iterations the global best position is selected as the best solution to the problem. The AIWPSO is proposed to balance the global exploration and local exploitation abilities for PSO. For each iteration, every particle chooses an appropriate inertia weight along the search space by dynamically adjusting the inertia weight $w$.

\item CS: Cuckoo Search~\cite{yang2009cuckoo, yang2010engineering} employs a combination of the L\'evy flight, which may be defined as a bird flight-inspired random walk with step $\tau$ over a Markov chain, together with a parasitic behavior of some cuckoo species. The model follows three basic ideas: i) each cuckoo lays one egg at a time in randomly chosen nests, ii) the host bird discover the cuckoo's egg with a probability $p_a\in[0, 1]$ and either discard the egg or abandon the chest and build a new one (a new solution is created), and iii) the nests with best eggs will carry over to the next generations.

\item FA: is derived from the fireflies' flash attractiveness when mating partners and attracting potential preys. Basically, the attractiveness $\beta$ of a firefly is computed by its position related to other fireflies in the swarm, as well as its brightness is determined by the value of the objective function at that position. Furthermore, the attractiveness depends on each firefly light absorption coefficient $\gamma$. In order to avoid local optima, the system is exposed to a random perturbation $\alpha$, and the best firefly performs a random walk across the search space.

\item BSA: it is a simple, effective and fast evolutionary algorithm developed to deal with problems characterized by slow computation and excessive sensitivity to control parameters. In a nutshell, it employs crossover and mutation operations together with a random selection of stored memories to generate a new population of individuals based on past experiences. BSA requires a proper selection of two parameters: the mixing rate ($mix\_rate$), which controls the number of elements of individuals that will mutate in the population, as well as the $F$ parameter, which controls the amplitude of the search-direction matrix.

\begin{sloppypar}
\item JADE: a differential evolution-based algorithm that implements the ``DE/current-to-$p$-best" mutation strategy, which employs only the $p$-best agents in the mutation process. Additionally, JADE uses an optional archive for historical information, as well as an adaptive updating in the control parameter. JADE requires the selection of the parameter $c$, which stands for the rate of parameter adaptation, and $g$ (greedness), that determines the greediness of the mutation strategy.
\end{sloppypar}

\item CoBiDE: it also a differential evolution-based technique that employs a covariance matrix for a better representation of the system's coordinates during the crossover process. Additionally, mutation and crossover are controlled using a bimodal distribution to achieve a good trade-off between exploration and exploitation. The probability to execute the differential evolution according to the covariance matrix is defined by the parameter $pb$, as well as the proportion of individuals chosen from the current population to calculate the covariance matrix is denoted by $ps$.

\end{itemize}

%% file: sections/methodology.tex
\section{Methodology}
\label{s.methodology}

This section provides a brief introduction to the concept of data reconstruction, as well as the description of the datasets and the experimental setup employed in this work.

\subsection{Data reconstruction}
\label{ss.reconstruction}

Although the literature is fulfilled with methods that employ image reconstruction for specific tasks, such as super-resolution image reconstruction~\cite{nguyen2001computationally,dong2014learning}, tomographies~\cite{liu1999optimization}, and denoising and debluring~\cite{puetter2005digital}, including RBM~\cite{pires2017robust} and DBM~\cite{pires2017deep} approaches, data reconstruction in the context of this paper stands for RBM-based models as an intrinsic process of the learning step, whose error is monitored for optimization purpose~\cite{hinton2012practical}, instead of a practical application itself. In other words, such models try to represent the input data in the hidden layers given their probability distribution. Such representation is supposedly capable of reconstructing a similar input given a stochastic Gibbs sampling. Afterward, the representation mentioned above is employed for a vast range of applications, such as classification~\cite{larochelle2008classification}, dimensionality reduction~\cite{hinton2006reducing}, modeling human motion~\cite{taylor2007modeling}, among others.

\subsection{Datasets}
\label{ss.datasets}

\begin{figure}[!ht]
  \centerline{\begin{tabular}{ccc}
      	\includegraphics[width=3.7cm,height=3.7cm]{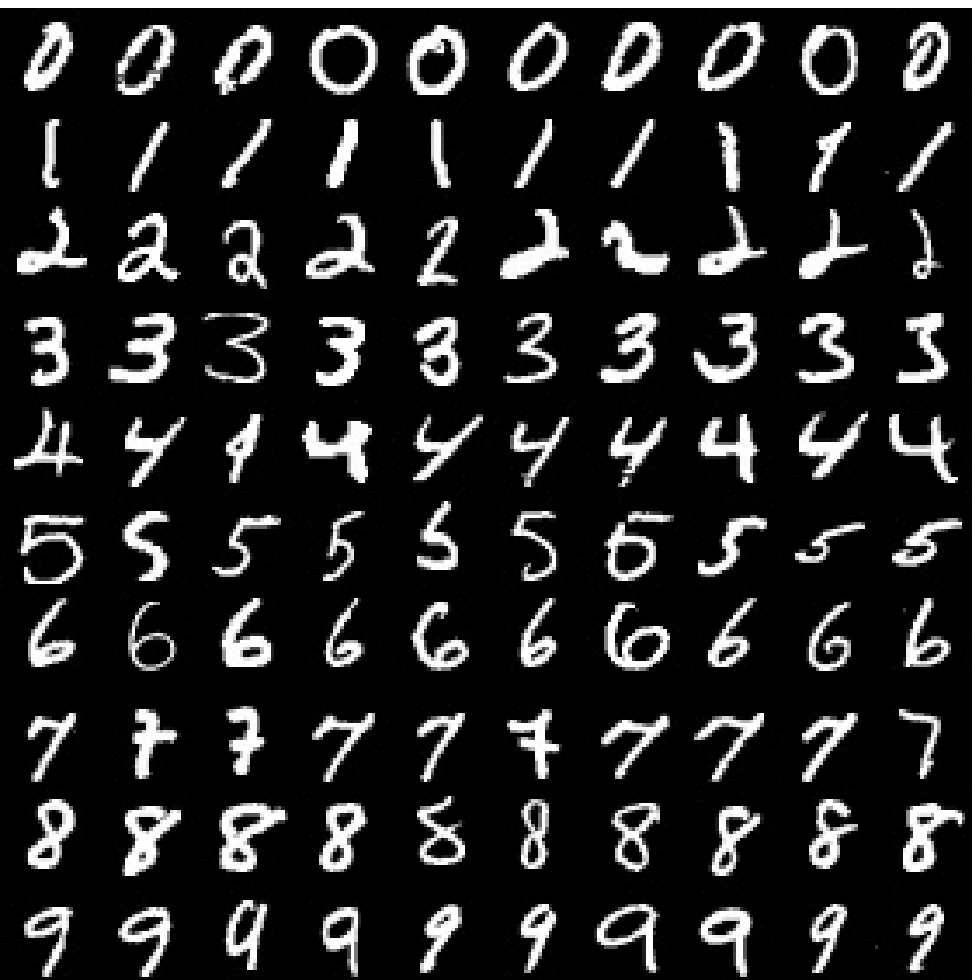} & 
      	\includegraphics[width=3.7cm,height=3.7cm]{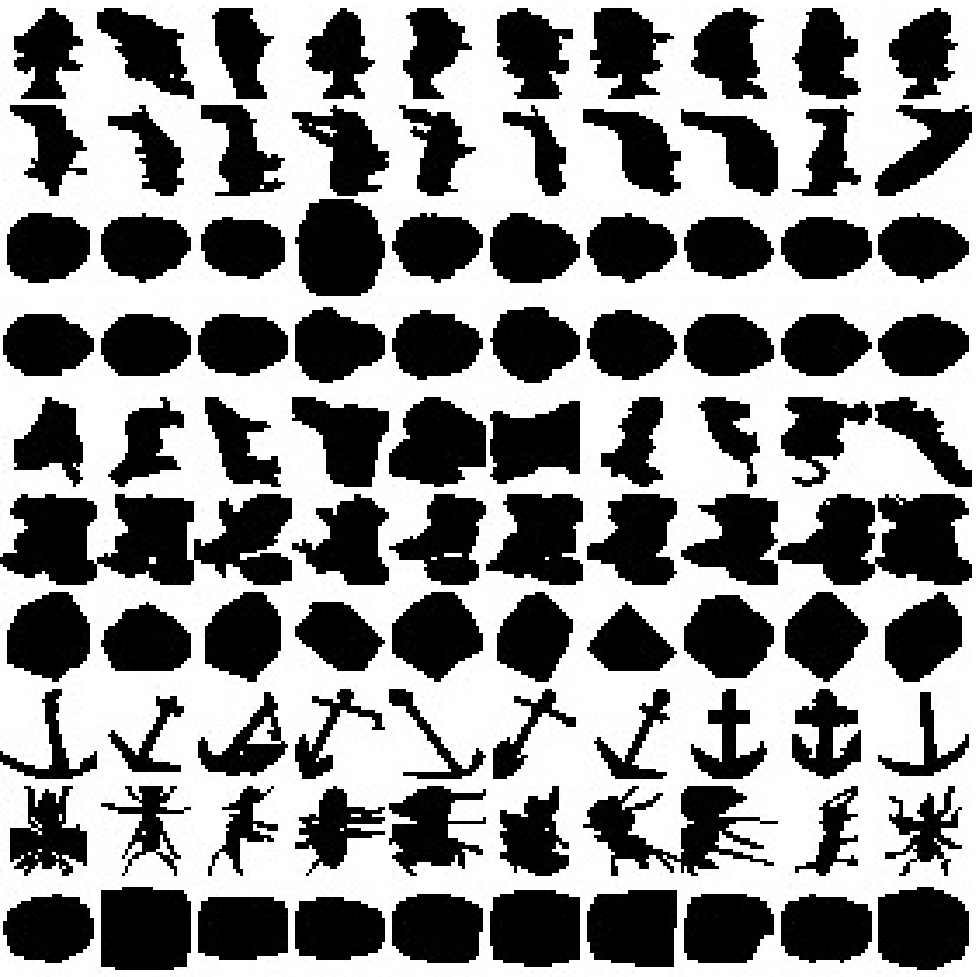} &
      	\includegraphics[width=3.7cm,height=3.7cm]{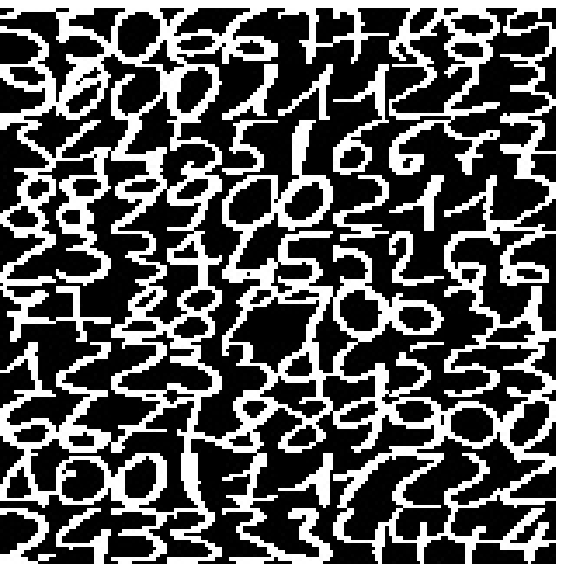} \\
      (a) & (b) & (c)
\end{tabular}}
\caption{Some training examples from (a) MNIST, (b) CalTech 101 Silhouettes and (c) Semeion datasets.}
\label{f.datasets}
\end{figure}

We validate DBM fine-tuning in the task of binary image reconstruction over three public datasets, as described below:

\begin{itemize}
\item MNIST dataset\footnote{\url{http://yann.lecun.com/exdb/mnist/}}: it is composed by images of handwritten digits. The original version contains a training set with $60,000$ images from digits `0'-`9', as well as a test set with $10,000$ images. Due to the high computational burden for RBM model selection, we decided to employ the original test set together with a reduced version of the training set\footnote{The original training set was reduced to $2\%$ of its former size, which corresponds to $1,200$ images.}.
\item Semeion Handwritten Digit Data Set\footnote{\url{https://archive.ics.uci.edu/ml/datasets/Semeion+Handwritten+Digit}}: composed of binary images of manuscript digits, this dataset contains 1,593 images with the resolution of $16\times16$ from around $80$ persons. The whole dataset was employed in the experimental section, being $2\%$ used for training purposes, as well as the remaining $98\%$ for testing.
\item CalTech 101 Silhouettes Data Set\footnote{\url{https://people.cs.umass.edu/~marlin/data.shtml}}: it is based on the former Caltech 101 dataset, and it comprises silhouettes of images from 101 classes with resolution of $28\times28$. We have used only the training and test sets, since our optimization model aims at minimizing the MSE error over the training set.
\end{itemize}
Figure~\ref{f.datasets} displays some training examples from both datasets, which were partitioned in 2\% for the training set and 98\% to compose the test set.

\subsection{Parameter Setting-up}
\label{ss.parameterSettingUp}

One of the main shortcoming in using RBM-based models, such as DBM and DBN, concerns their fine-tuning hyperparameter task, which aims at selecting a suitable set of parameters in such a way that the reconstruction error is minimized. In this work, we considered IHS, FA, CS, AIWPSO, BSA, JADE, and the CoBiDE against RS for DBM hyperparameter fine-tuning. We also evaluated the robustness of the proposed approach using three distinct DBN and DBM models: one layer (1L), two layers (2L) and three layers (3L). Finally, Table~\ref{t.parameters} presents the parameters used for each optimization technique\footnote{Parameters were empirically selected based on each technique author's suggestions, as well as the values commonly adopted in the literature~\cite{papaQUATERNION:17, rosa2016learning, PapaGECCO:15, RodriguesBook:16, passosiRBM:2017}}, where $5$ agents (initial solutions) were used for all optimization techniques during $50$ iterations for convergence~\footnote{The selected number of agents and iterations for convergence were empirically chosen based on values commonly adopted in the literature~\cite{papaQUATERNION:17, rosa2016learning, PapaGECCO:15}}.. 

\begin{table}[h]
\centering
\caption{Parameter configuration.}
\begin{tabular}{c|c} \hline
Technique&Parameters\\ \hline\hline
IHS & $HMCR=0.7$, $PAR_{MIN}=0.1$ \\ 
    & $PAR_{MAX}=0.7$, $\varrho_{MIN}=1$\\ 
    & $\varrho_{MAX}=10$\\ \hline
AIWPSO & $c_1=1.7$, $c_2=1.7$\\ 
	& $w=0.7, w_{MIN}=0.5, w_{MAX}=1.5$ \\ \hline
CS & $\tau=0.1$, $\tau_{MIN}=0.5$, $\tau_{MAX}=1$\\
   	& $p=0.25$, $p_{MIN}=0.05$, $p_{MAX}=0.5$\\ \hline
FA & $\gamma=1$, $\beta=1$, $\alpha=0.2$\\ \hline
BSA & $mix\_rate=1.0$, $F=3$\\ \hline
JADE & $c=0.1$, $g=0.05$\\ \hline
CoBiDE & $pb=0.4$, $ps=0.5$\\ \hline
\end{tabular}
\label{t.parameters}
\end{table}

We conducted a cross-validation approach with $20$ runnings, $10$ iterations for the learning procedure of each RBM, and mini-batches of size $20$. In addition, we also considered two learning algorithms: Contrastive Divergence (CD)~\cite{Hinton:02} and Persistent Contrastive Divergence (PCD)~\cite{TielemanICML:08}. Finally, the Wilcoxon signed-rank test~\cite{Wilcoxon:45} with significance of $0.05$ was used for statistical validation purposes.

Finally, the codes used to reproduce the experiments of the paper are available on GitHub\footnote{LibOPF: https://github.com/jppbsi/LibOPF}\footnote{LibDEEP: https://github.com/jppbsi/LibDEEP}\footnote{LibDEV: https://github.com/jppbsi/LibDEV}\footnote{LibOPT~\cite{PapaLIBOPT:17}: https://github.com/jppbsi/LibOPT}. The experiments were conducted using an Ubuntu 16.04 Linux machine with 64Gb of RAM running an $2$x Intel$^{\textregistered}$ Xeon Bronze $3106$ with a frequency of $1.70$ GHz. All the coding was built in C.

%% file: sections/experiments.tex
\section{Experiments}
\label{s.experiments}

In this section, we present the experimental results concerning DBM and DBN hyperparameter optimization on the task of binary image reconstruction. Both techniques were compared using two different learning algorithms, i.e. Contrastive Divergence and Persistent Contrastive Divergence. Also, seven optimization methods were employed. Additionally, three distinct models used for comparison purposes: one layer (1L), two layers (2L), and three (3L) layers. 

\subsection{Experimental Results}
\label{ss.experimentalResults}

Tables~\ref{t.results_mnist} presents the average values of the minimum squared error over the MNIST dataset, being the values in bold the best results considering the Wilcoxon signed-rank test. One can observe the metaheuristic techniques obtained the best results, with special attention to IHS, JADE, and CoBiDE for both DBN and DBM models. Also, one can not figure a considerable difference between shallow and deep models, since we limited the number of iterations for convergence to $10$, as well as we did not employ fine-tuning as a final step for DBN and DBM connection weights. The main reasons for limiting the number of iterations are related to time constraints, as well as the convergence process itself. As a matter of fact, if one has unlimited resources in terms of computational load, a standard random search may obtain results as good as the ones obtained by metaheuristic techniques, since they will have enough time for convergence. However, we would like to emphasize that DBM hyperparameter fine-tuning is quite useful when time is limited and a serious constraint.

\begin{table}[!htb]
\caption{Average MSE values and standard deviation considering MNIST dataset.}
\begin{center}
\resizebox{\columnwidth}{!}{%
\begin{tabular}{|l|l|c|c|c|c|c|c|c|c|c|c|c|c|}
\cline{3-14}
     \multicolumn{2}{c|}{}   & \multicolumn{4}{c|}{1L} & \multicolumn{4}{c|}{2L} & \multicolumn{4}{c|}{3L} \\ \cline{3-14} 
     \multicolumn{2}{c|}{} 	  & \multicolumn{2}{c|}{DBN} & \multicolumn{2}{c|}{DBM} & \multicolumn{2}{c|}{DBN} & \multicolumn{2}{c|}{DBM}  & \multicolumn{2}{c|}{DBN} & \multicolumn{2}{c|}{DBM}   \\ \cline{3-14} 
\multicolumn{1}{c}{Technique}&  \multicolumn{1}{c|}{Statistics} & CD    & PCD    & CD   & PCD   & CD    & PCD    & CD   & PCD   & CD    & PCD    & CD   & PCD   \\ \hline
\multirow{2}{*}{ {IHS}}  & Mean &\textbf{0.08758} & 0.08762 & \textbf{0.08744} & 0.08766 & 0.08762 & 0.08762 & 0.08761 & 0.08761 & 0.08762 & 0.08762 & 0.08760 & 0.08761 \\ 
	& Std. & 8.102e-05 & 7.581e-05  &  3.702e-04 &  4.686e-04  &  5.203e-05  &  6.018e-05  &  5.063e-05  & 3.834e-05  &  6.971e-05  &  5.941e-05 &  5.845e-05  &  5.885e-05  \\\hline
\multirow{2}{*}{ {AIWPSO}}  & Mean & 0.08764 & 0.08761 & 0.08765 & 0.08771 & 0.08763 & 0.08762 & 0.08762 & 0.08761 & 0.08762 & 0.08762 & \textbf{0.08759} & 0.08760  \\ 
		& Std. &5.694e-05 & 4.728e-05  & 4.793e-04 &  3.744e-04  &  5.965e-05  &  4.879e-05  &  5.207e-05  &  5.250e-05  &  4.299e-05  &  4.643e-05 &  5.280e-05  &  5.505e-05  \\\hline
\multirow{2}{*}{ {CS}}  & Mean &0.08763 & 0.08764 & 0.08767 & 0.08770 & 0.08764 & 0.08765 & 0.08760 & 0.08760 & 0.08764 & 0.08765 & 0.08762 & 0.08761  \\ 
		& Std. & 5.393e-05 &6.722e-05  &  7.988e-05 &  2.713e-04  &6.906e-05  &  5.766e-05  &  5.771e-05  & 6.122e-05  &  6.611e-05  & 8.424e-05 &  5.541e-05  & 5.356e-05 \\\hline
\multirow{2}{*}{ {FA}} &Mean & 0.08763 & 0.08764 & 0.08766 & 0.08762 & 0.08763 & 0.08763 & 0.08761 & 0.08763 & 0.08763  & 0.08763 & 0.08761 & 0.08761  \\ 
		& Std. & 6.749e-05 & 6.271e-05  &  1.113e-04 &  2.673e-04  &  5.923e-05  & 6.488e-05  &  4.780e-05  &  8.191e-05  &  6.342e-05  &  5.658e-05&  3.951e-05 &  6.131e-05  \\\hline
\multirow{2}{*}{ {BSA}} & Mean &0.08762 & 0.08762 & 0.08774 & 0.08766 & 0.08762 & 0.08763 & 0.08761 & 0.08762 & 0.08763 & 0.08762 & 0.08762 & 0.08762  \\ 
		& Std. & 5.231e-05 & 6.697e-05  &  4.135e-04 & 3.242e-04  &  6.697e-05 &  6.555e-05  &  4.072e-05  & 5.870e-05  & 6.176e-05  &  6.785e-05 &  5.416e-05  & 5.175e-05 \\\hline
\multirow{2}{*}{ {JADE}} & Mean &0.08760 & 0.08763 & \textbf{0.08754} & \textbf{0.08749} &0.08763 & 0.08764 & 0.08761 & 0.08761 & 0.08763 & 0.08763 & 0.08761 &0.08761  \\ 
		& Std. & 6.780e-05 & 5.644e-05  & 4.131e-04 & 3.256e-04  & 6.264e-05  &5.967e-05  & 6.284e-05  & 5.491e-05 &  6.546e-05  & 6.696e-05 & 5.662e-05  & 5.356e-05  \\\hline
\multirow{2}{*}{ {CoBiDE}}   & Mean &0.08763 & 0.08762 & \textbf{0.08757} & 0.08765 & 0.08763 & 0.08764 & 0.08762 & 0.08760 & 0.08763  & 0.08762 & 0.08761 & 0.08760  \\ 
		& Std. & 6.249e-05 & 7.203e-05  & 4.104e-04 & 3.460e-04  & 6.053e-05  & 5.312e-05 & 6.786e-05  & 5.359e-05 & 6.022e-05  & 6.219e-05 & 5.222e-05 & 4.868e-05 \\\hline
\multirow{2}{*}{ {RS}} & Mean &0.08762 & 0.08763 & 0.08780 & 0.08782 & 0.08762 & 0.08763 & 0.08761 & 0.08760 & 0.08763 & 0.08763 & 0.08761 & 0.08761  \\ 
		& Std. & 5.699e-05 & 5.495e-05  &  3.965e-04 &  5.091e-04  &  4.355e-05  &  4.765e-05  &  4.657e-05  & 5.008e-05 &  6.911e-05  &  5.740e-05 &  5.125e-05  & 5.979e-05 \\\hline
\end{tabular}}
\label{t.results_mnist}
\end{center}
\end{table}

Table~\ref{t.results_caltech} presents the results concerning CalTech 101 Silhouettes dataset. In this case, the best results were achieved by DBN with one layer only. Caltech poses a greater challenge, since it has more classes than MNIST, which should us to believe more iterations for convergence would be required for DBM learning, since it a more complex model than DBN. Also, the best results were obtained by means of Improved Harmony Search, BSA, JADE, and CoBiDE.

\begin{table}[!htb]
\caption{Average MSE values and standard deviation considering CalTech 101 Silhouettes dataset.}
\begin{center}
\resizebox{\columnwidth}{!}{%
\begin{tabular}{|l|l|c|c|c|c|c|c|c|c|c|c|c|c|}
\cline{3-14}
     \multicolumn{2}{c|}{}   & \multicolumn{4}{c|}{1L} & \multicolumn{4}{c|}{2L} & \multicolumn{4}{c|}{3L} \\ \cline{3-14} 
     \multicolumn{2}{c|}{} 	  & \multicolumn{2}{c|}{DBN} & \multicolumn{2}{c|}{DBM} & \multicolumn{2}{c|}{DBN} & \multicolumn{2}{c|}{DBM}  & \multicolumn{2}{c|}{DBN} & \multicolumn{2}{c|}{DBM}   \\ \cline{3-14} 
\multicolumn{1}{c}{Technique}&  \multicolumn{1}{c|}{Statistics} & CD    & PCD    & CD   & PCD   & CD    & PCD    & CD   & PCD   & CD    & PCD    & CD   & PCD   \\ \hline
\multirow{2}{*}{ {IHS}}  & Mean&\textbf{0.15554} & 0.15731 & 0.15983 & 0.15980 & 0.16057 & 0.16054 & 0.16055 & 0.16055 & 0.16059 & 0.16058 & 0.16057 & 0.16056\\ 
		&  Std. &  2.107e-03 &  1.584e-03  &   1.064e-03 &   7.218e-04  &  1.980e-04  &   2.922e-04  &   1.958e-04  &   1.852e-04  &   2.162e-04  &   2.078-04 &   2.041e-04  &   2.150e-04  \\\hline
\multirow{2}{*}{ {AIWPSO}}  & Mean & 0.15641 & 0.15825 & 0.16006 & 0.16014 & 0.16056 & 0.16060 & 0.16056 & 0.16061 & 0.16058 & 0.16057 & 0.16057 & 0.16057  \\ 
		&  Std. &  2.414e-03 &  2.310e-03  &   8.199e-04 &   7.570e-04  &  2.010e-04  &   2.224e-04  &   1.914e-04  &   2.291e-04  &   2.192e-04  &   2.129e-04 &   1.890e-04  &   2.124e-04  \\\hline
\multirow{2}{*}{ {CS}}  & Mean & 0.15923 & 0.15992 & 0.16023 & 0.16024 & 0.16057 & 0.16062 & 0.16057 & 0.16056 & 0.16059 & 0.16061 & 0.16055 & 0.16057  \\ 
		&  Std. &  1.707e-03 &  1.030e-03  &   4.329e-04 &   3.538e-04  &  1.855e-04  &   2.275e-04  &   2.071e-04  &   2.107e-04  &   2.123e-04  &   2.034e-04 &   1.941e-04  &   2.123e-04  \\\hline
\multirow{2}{*}{ {FA}} &Mean & 0.16002 & 0.15956 & 0.16051 & 0.16034 & 0.16060 & 0.16058 & 0.16069 & 0.16056 &  0.16060 & 0.16058 & 0.16055 & 0.16055  \\ 
		&  Std. &  1.555e-03 &  1.176e-03  &   5.541e-04 &   6.887e-04  &   2.120e-04  &   2.130e-04  &   6.536e-04  &   2.147e-04  &   2.327e-04  &   2.098e-04 &   2.174e-04  &  2.029e-04  \\\hline
\multirow{2}{*}{ {BSA}} & Mean & \textbf{0.15599} & 0.15775 & 0.15992 & 0.15983 & 0.16056 &0.16056 & 0.16052 & 0.16054 & 0.16057  & 0.16058 & 0.16057 & 0.16055 \\ 
		&  Std. &  1.542e-03 &  1.511e-03  &   8.302e-03 &   6.978e-04 &   2.016e-04  &   2.174e-04  &   1.770e-04  &   1.985e-04 &   2.063e-04  &   1.981e-04 &   1.878e-04  &   2.004e-04  \\\hline
\multirow{2}{*}{ {JADE}} & Mean & \textbf{0.15608} & 0.15790 & 0.15945 & 0.15988 & 0.16058 & 0.16057 & 0.16055 & 0.16058 & 0.16059  & 0.16057 & 0.16058 & 0.16054  \\ 
		&  Std. &  1.835e-03 &  1.351e-03  &   6.426e-04 &   6.015e-04  &   2.037e-04  &   2.001e-04  &   1.876e-04  &   1.784e-04  &   1.933e-04  &   2.131e-04 &   2.126e-04  &   2.000e-04  \\\hline
\multirow{2}{*}{ {CoBiDE}}   & Mean & \textbf{0.15638} & 0.15800 & 0.15982 & 0.15982 & 0.16059 & 0.16057 & 0.16059 & 0.16056 & 0.16060  & 0.16059 & 0.16056 & 0.16054  \\ 
		&  Std. &  1.912e-03 &  1.209e-03  &   6.181e-04 &   8.848e-04  &   2.298e-04  &   2.204e-04  &   3.093e-04 &   1.652e-04  &   2.090e-04  &   2.023e-04 &   1.739e-04  &   2.060e-04  \\\hline
\multirow{2}{*}{ {RS}} & Mean & 0.15676 & 0.15845 & 0.15967 & 0.15976 & 0.16060 & 0.16062 &  0.16059 & 0.16057 & 0.16057 & 0.16056 & 0.16056 & 0.16056 \\ 
		&  Std. &  1.623e-03 &  1.220e-03  &   7.164e-04 &   7.133e-04  &   1.998e-04  &   1.915e-04  &   1.974e-04  &   1.993e-04  &   1.998e-04  &   2.173e-04 &   1.853e-04  &   1.868e-04  \\\hline
\end{tabular}}
\label{t.results_caltech}
\end{center}
\end{table}

\begin{sloppypar}
Table~\ref{t.results_semeion} presents the results obtained over Semeion Handwritten Digit dataset, being IHS and JADE the most accurate techniques. The best results concerning MNIST and Semeion Handwritten Digits datasets, as can be clearly seen on Tables~\ref{t.results_mnist} and ~\ref{t.results_semeion}, was acquired using the DBM. DBN, however, had the best results considering CalTech 101 Silhouettes dataset, as presented in Table~\ref{t.results_caltech}. Some interesting conclusions can be extracted from a closer look at these results:  (i) meta-heuristic-based optimization allows more accurate results than a random search, as argued by the works of Papa et al.~\cite{PapaGECCO:15,PapaJoCS:15,PapaASC:15} already; (ii) DBMs seem to produce more accurate results than DBNs; (iii) the number of layers do not seem to influence the results when one fine-tune parameters; (iv) IHS achieved the best results in all datasets (concerning both DBN and DBN), but with results statistically similar to other meta-heuristic techniques as well; and (v) we could not realize a significant difference between CD and PCD, since we employed $10$ iterations for learning only. Actually, PCD is expected to work better, but at the price of a longer convergence process.
\end{sloppypar}

\begin{table}[!htb]
\caption{Average MSE values and standard deviation considering Semeion Handwritten Digit dataset.}
\begin{center}
\resizebox{\columnwidth}{!}{%
\begin{tabular}{|l|l|c|c|c|c|c|c|c|c|c|c|c|c|}
\cline{3-14}
     \multicolumn{2}{c|}{}   & \multicolumn{4}{c|}{1L} & \multicolumn{4}{c|}{2L} & \multicolumn{4}{c|}{3L} \\ \cline{3-14} 
     \multicolumn{2}{c|}{} 	  & \multicolumn{2}{c|}{DBN} & \multicolumn{2}{c|}{DBM} & \multicolumn{2}{c|}{DBN} & \multicolumn{2}{c|}{DBM}  & \multicolumn{2}{c|}{DBN} & \multicolumn{2}{c|}{DBM}   \\ \cline{3-14} 
\multicolumn{1}{c}{Technique}&  \multicolumn{1}{c|}{Statistics} & CD    & PCD    & CD   & PCD   & CD    & PCD    & CD   & PCD   & CD    & PCD    & CD   & PCD   \\ \hline
\multirow{2}{*}{ {IHS}}  & Mean& 0.19359 & 0.20009 & \textbf{0.19025} & \textbf{0.19078} & 0.20961 & 0.20961 & 0.20956 & 0.20956 & 0.20961 & 0.20963 & 0.20958 & 0.20958\\ 
		&  Std. &  1.367e-03 &  1.965e-03  &   8.901e-04 &   1.367e-03  &   3.669e-04  &   3.637e-04  &   3.571e-04  &   3.438e-04  &   3.731e-04  &   3.772e-04 &   3.609e-04  &   3.417e-04  \\\hline
\multirow{2}{*}{ {AIWPSO}}  & Mean & 0.20044 & 0.20274 & 0.19679 & 0.19426 & 0.20959 & 0.20961 & 0.20958 & 0.20956 & 0.20964 & 0.20961 & 0.20959 & 0.20959  \\ 
		&  Std. &  6.856-03 &  3.994  &   7.995e-03 &   7.044e-03  &   3.521e-04  &   3.853e-04  &   3.644e-04  &   3.619e-04  &   3.773e-04  &   3.584e-04 &   3.784e-04  &   3.664e-04  \\\hline
\multirow{2}{*}{ {CS}}  & Mean & 0.20528 & 0.20647 & 0.20728 & 0.20651 & 0.20965 & 0.20960 & 0.20957 & 0.20959 & 0.20964 & 0.20963 & 0.20960 & 0.20960 \\ 
		&  Std. &  4.948-03 &  3.556e-03 &   2.894e-03 &   2.352e-03  &   4.034e-04  &   3.554e-04  &   3.616e-04  &   3.722e-04  &  3 .696e-04  &   3.572e-04 &   3.612e-04  &   3.430e-04  \\\hline
\multirow{2}{*}{ {FA}} &Mean & 0.20638 & 0.20894 & 0.20649 & 0.20319 & 0.20966 & 0.20965 & 0.20960 & 0.20960 & 0.20964 & 0.20965 & 0.20960 & 0.20928 \\ 
		&  Std. &  4.922-03 &  2.085e-03  &   5.630e-03 &   7.548e-03  &   4.098e-04  &   3.855e-04  &   3.609e-04  &   3.605e-04  &   3.499e-04  &   4.117e-04 &   3.387e-04  &   1.555e-03  \\\hline
\multirow{2}{*}{ {BSA}} & Mean & 0.19571 & 0.20002  & 0.19221 & 0.19325  & 0.20961 & 0.20959 & 0.20960 & 0.20958 & 0.20962  & 0.20962 & 0.20960 & 0.20956  \\ 
		&  Std. &  3.648-03 &  2.544e-03  &   2.879e-03 &   2.419e-03  &   3.591e-04  &   3.683e-04  &   3.783e-04  &   3.480e-04  &   3.722e-04  &   3.847e-04 &   3.716e-04  &   3.600e-04  \\\hline
\multirow{2}{*}{ {JADE}} & Mean & 0.19893 & 0.20165 & \textbf{0.19152} & \textbf{0.19170} & 0.20962 & 0.20960 & 0.20957 & 0.20958 & 0.20964  & 0.20959 & 0.20956 & 0.20961  \\ 
		&  Std. &  7.890-03 &  5.316e-03  &   4.213e-03 &   4.410e-03  &   3.554e-04  &   3.602e-04  &   3.501e-04  &   3.755e-04  &   3.579e-04  &   3.708e-04 &   3.524e-04  &   3.899e-04  \\\hline
\multirow{2}{*}{ {CoBiDE}}   & Mean & 0.19328 & 0.19896 & 0.19190 & 0.19138 & 0.20962 & 0.20961 & 0.20959 & 0.20958 & 0.20960  & 0.20961 & 0.20958 & 0.20959  \\ 
		&  Std. &  1.332-03 &  1.478e-03  &  1.821e-03 &   1.556e-03  &   3.505e-04  &   3.631e-04  &   3.678e-04  &   3.550e-04  &   3.579e-04  &   4.119e-04 &   3.664e-04  &   3.593e-04  \\\hline
\multirow{2}{*}{ {RS}} & Mean & 0.19710 & 0.20361 & 0.19458 & 0.19463 & 0.20962 & 0.20959 & 0.20960 & 0.20957 & 0.20960 & 0.20960 & 0.20960 & 0.20959  \\ 
		&  Std. &  3.133e-03 &  1.837e-03  &   3.891e-03 &   3.909e-03  &   3.621e-04  &   3.494e-04  &   3.864e-04  &   3.680e-04  &   3.677e-04  &   3.563e-04 &   3.538e-04  &   3.410e-04  \\\hline
\end{tabular}}
\label{t.results_semeion}
\end{center}
\end{table}

\begin{figure}[ht]
  \centerline{
    \begin{tabular}{cc}
	\includegraphics[width=6.5cm]{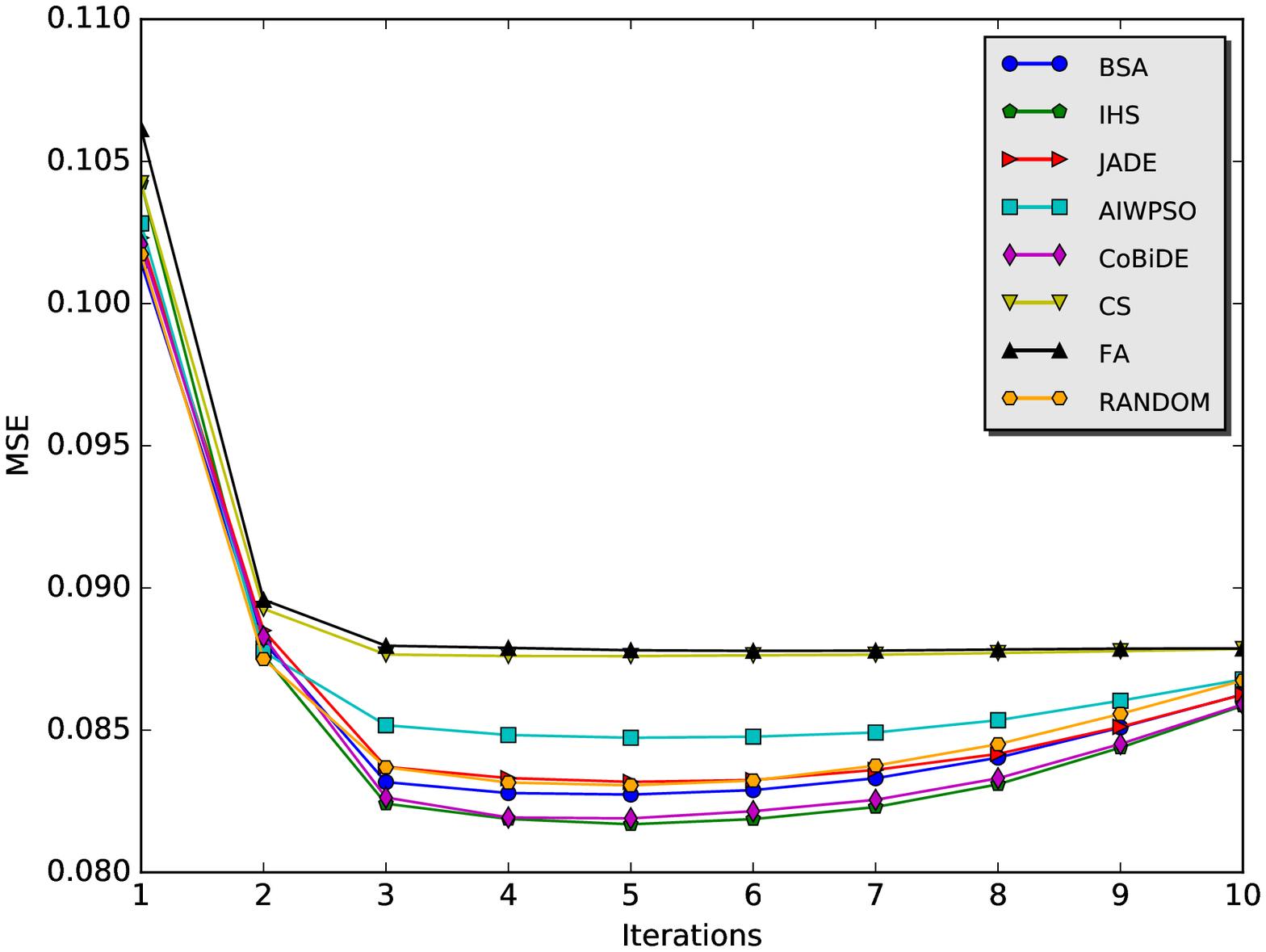} &
	\includegraphics[width=6.5cm]{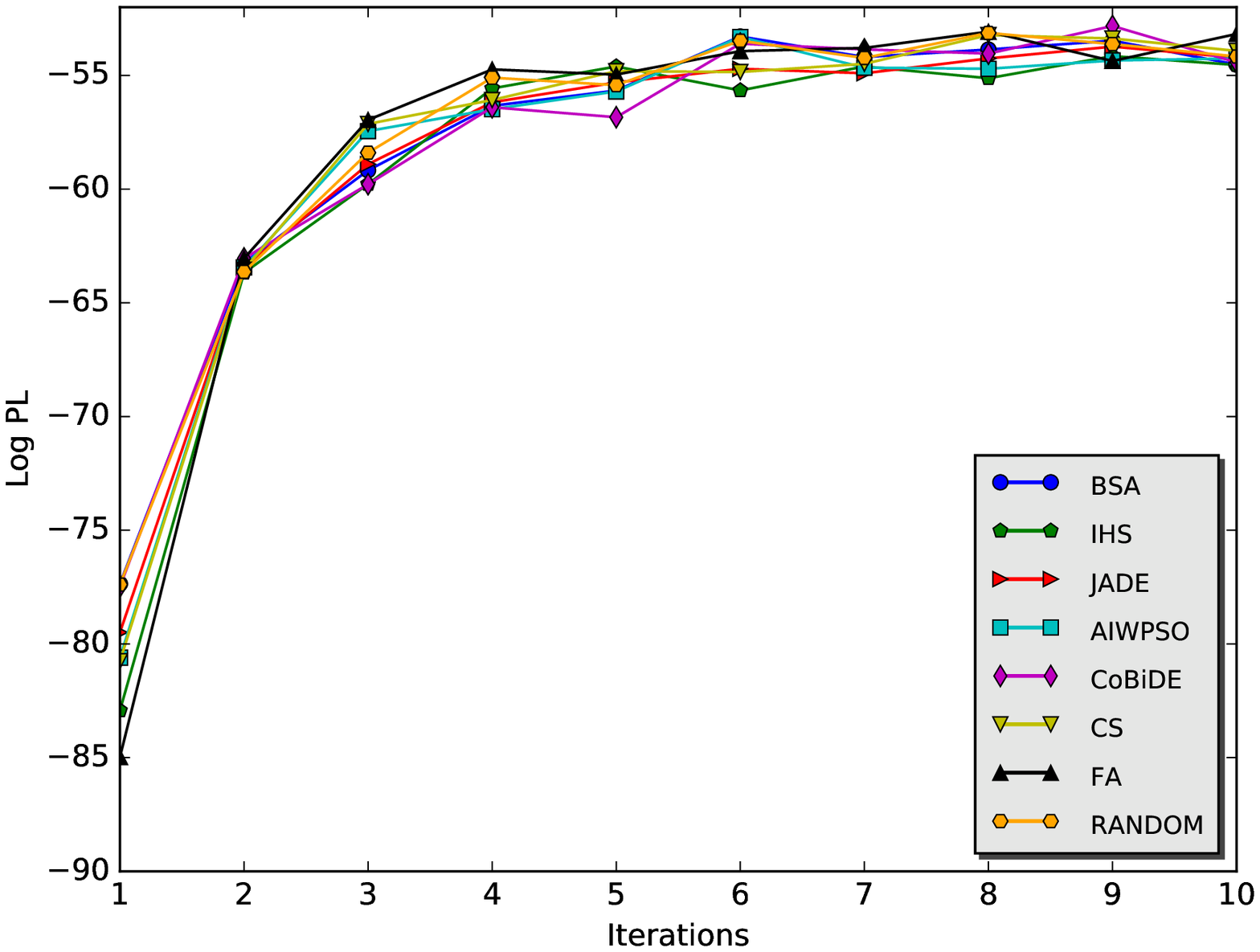}\\
	(a) & (b)
    \end{tabular}}
    \caption{\label{f.evolutionDBM}MSE and Log PL values during the convergence process considering DBM over MNIST dataset for (a) and (b), respectively.}
\end{figure}

\begin{figure}[ht]
  \centerline{
    \begin{tabular}{cc}
	\includegraphics[width=6.5cm]{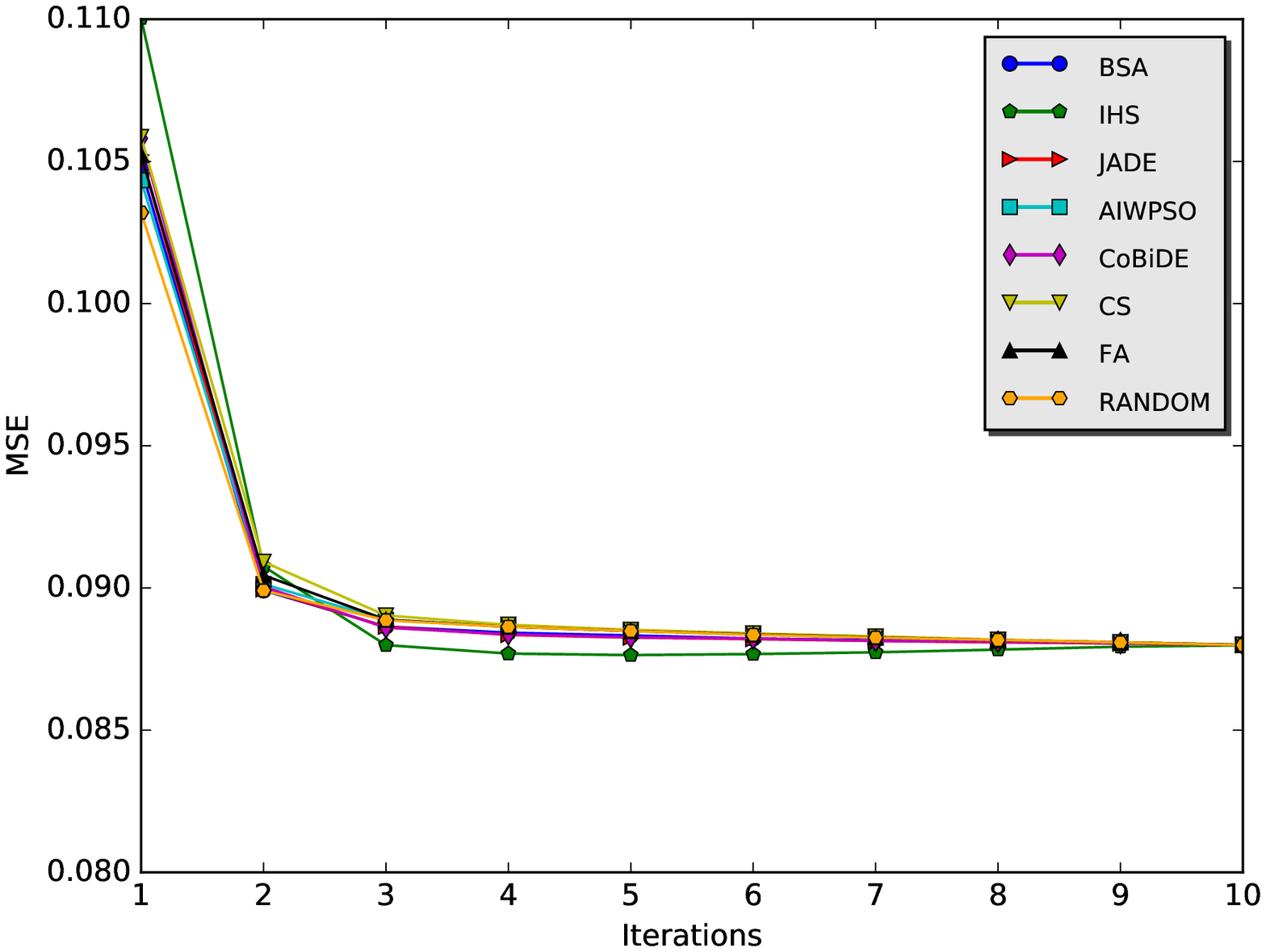} &
	\includegraphics[width=6.5cm]{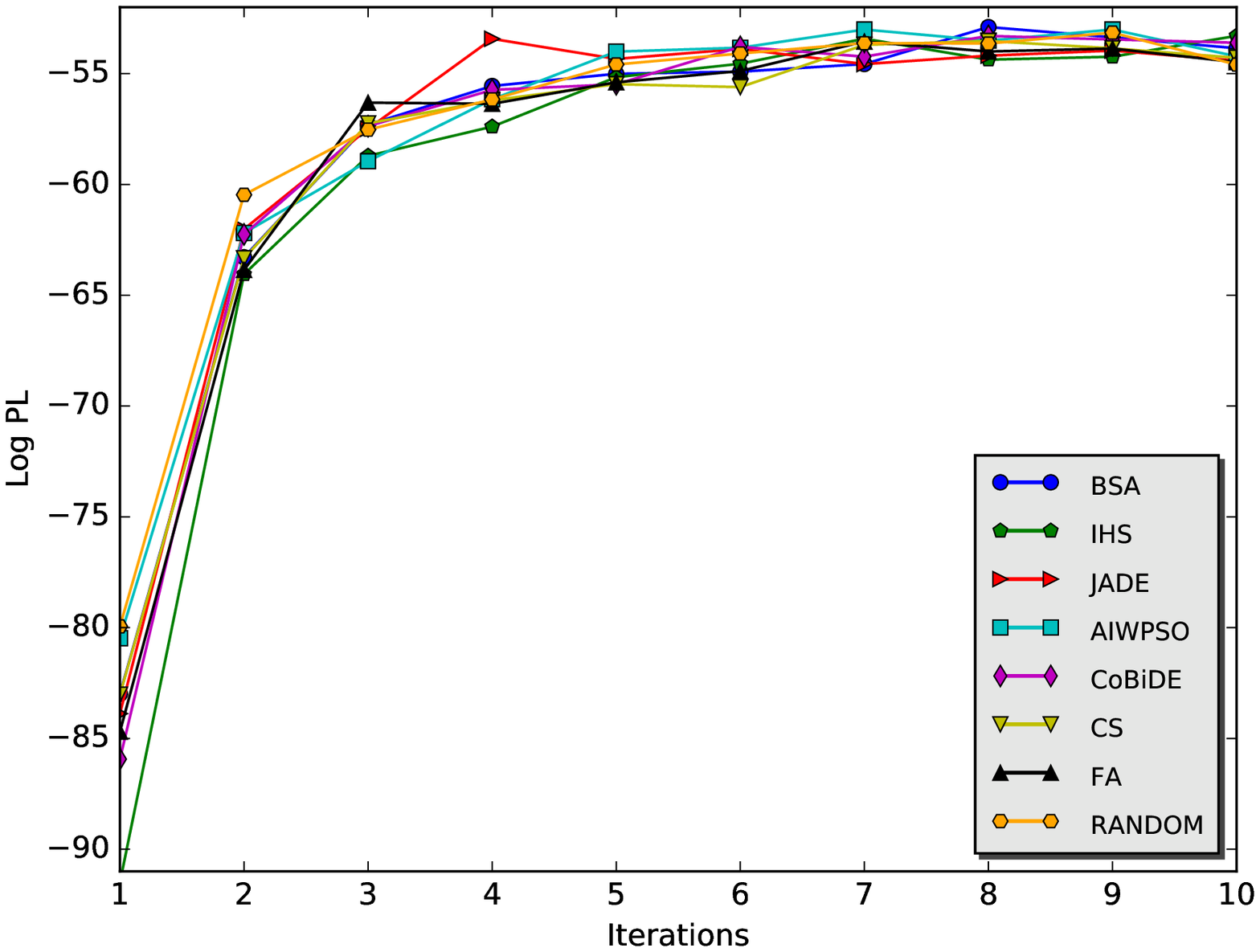}\\
	(a) & (b)
    \end{tabular}}
    \caption{\label{f.evolutionDBN}MSE and Log PL values during the convergence process considering DBN over MNIST dataset for (a) and (b), respectively.}
\end{figure}

Figures~\ref{f.evolutionDBM} and~\ref{f.evolutionDBN} display the convergence process regarding the mean squared error (MSM) and logarithm of the pseudo-likelihood (PL) values obtained during the learning step for DBM and DBN, respectively,  trained with CD over MNIST dataset. We used the mean values of the first layer for all optimization algorithms. One can observe DBM obtained the better approximation of the model during all iterations, and both ended up with similar log PL values (iteration \#10). However, it is important to shed light over the main contribution of this paper is not to show DBM may learn better models than DBNs, but to stress meta-heuristic techniques are suitable to fine-tune DBM parameters as well. 

Although one can realize an oscillating behavior of the optimization techniques, all of them obtained better models at the last iteration (i.e. a highest log PL) than RS, except for the nature-inspired algorithms, that achieved similar results in most of the experiments, probably due to its demand for more iterations to convergence. The results implies that using meta-heuristic techniques to fine-tune DBMs seems to be reasonable. DBMs optimized by meta-heuristic-based techniques obtained the best results considering all datasets used in this work as well.

\subsection{Statistical Analysis}
\label{ss.statisticalAnalisys}

In this section, we detailed the Wilcoxon signed-rank test obtained through a pairwise comparison among the techniques. For such purpose, we used $5\%$ of significance to provide the statistical similarity among the best results obtained by each technique, i.e., considering both number of layers and learning algorithm. Tables~\ref{t.statistical_mnist},~\ref{t.statistical_caltech} and~\ref{t.statistical_semeion} presents the statistical evaluation concerning MNIST, CalTech 101 Silhouettes, and Semeion datasets. 

\begin{table}[!htb]
\caption{Statistical analysis considering MNIST dataset.}
\begin{center}
\resizebox{\columnwidth}{!}{%
\begin{tabular}{l| c c c c c c c c }
\cline{2-9}
                          & IHS    & AIWPSO    & CS   & FA   & BSA    & JADE    & CoBiDE   &  RS     \\ \hline
 IHS  &   \cellcolor[gray]{0.8} & \cellcolor[gray]{0.8} &     \cellcolor[gray]{0.8} & \cellcolor[gray]{0.8} & \cellcolor[gray]{0.8} & \cellcolor[gray]{0.8} &  \cellcolor[gray]{0.8} & \cellcolor[gray]{0.8} \\ 
 AIWPSO  & $=$  & \cellcolor[gray]{0.8}  &     \cellcolor[gray]{0.8} & \cellcolor[gray]{0.8} & \cellcolor[gray]{0.8} & \cellcolor[gray]{0.8} &  \cellcolor[gray]{0.8} & \cellcolor[gray]{0.8}   \\ 
CS &  $\neq$   & $=$   &   \cellcolor[gray]{0.8} & \cellcolor[gray]{0.8} & \cellcolor[gray]{0.8} & \cellcolor[gray]{0.8} &  \cellcolor[gray]{0.8} & \cellcolor[gray]{0.8} \\
FA  & $\neq$   &  $=$   &  $=$  &    \cellcolor[gray]{0.8}  & \cellcolor[gray]{0.8} & \cellcolor[gray]{0.8} &  \cellcolor[gray]{0.8} & \cellcolor[gray]{0.8} \\
BSA  &  $\neq$   &  $\neq$ &   $=$    &  $=$   & \cellcolor[gray]{0.8} & \cellcolor[gray]{0.8} &  \cellcolor[gray]{0.8} & \cellcolor[gray]{0.8}  \\
JADE  &  $=$  & $=$  &   $=$   & $=$  &  $=$   & \cellcolor[gray]{0.8} &  \cellcolor[gray]{0.8} & \cellcolor[gray]{0.8}   \\ 
CoBiDE  & $=$  &  $=$  & $=$   & $=$  &  $=$  &  $=$ &  \cellcolor[gray]{0.8} & \cellcolor[gray]{0.8} \\ 
RS  & $\neq$   & $=$   &  $=$  & $=$  & $=$ & $=$  & $=$ &  \cellcolor[gray]{0.8}  \\ \hline

\end{tabular}}
\label{t.statistical_mnist}
\end{center}
\end{table}

\begin{table}[!htb]
\caption{Statistical analysis considering CalTech 101 Silhouettes dataset.}
\begin{center}
\resizebox{\columnwidth}{!}{%
\begin{tabular}{l| c c c c c c c c }
\cline{2-9}
                          & IHS    & AIWPSO    & CS   & FA   & BSA    & JADE    & CoBiDE    & RS     \\ \hline
 IHS  &   \cellcolor[gray]{0.8} & \cellcolor[gray]{0.8} &     \cellcolor[gray]{0.8} & \cellcolor[gray]{0.8} & \cellcolor[gray]{0.8} & \cellcolor[gray]{0.8} &  \cellcolor[gray]{0.8} & \cellcolor[gray]{0.8} \\ 
 AIWPSO  & $\neq$   & \cellcolor[gray]{0.8}  &     \cellcolor[gray]{0.8} & \cellcolor[gray]{0.8} & \cellcolor[gray]{0.8} & \cellcolor[gray]{0.8} &  \cellcolor[gray]{0.8} &\cellcolor[gray]{0.8}  \\ 
CS &  $\neq$    &   $\neq$  &   \cellcolor[gray]{0.8} & \cellcolor[gray]{0.8} & \cellcolor[gray]{0.8} & \cellcolor[gray]{0.8} &  \cellcolor[gray]{0.8} & \cellcolor[gray]{0.8} \\
FA  & $\neq$   & $\neq$  &  $=$  &    \cellcolor[gray]{0.8}  & \cellcolor[gray]{0.8} & \cellcolor[gray]{0.8} &  \cellcolor[gray]{0.8} & \cellcolor[gray]{0.8}  \\
BSA  & $=$   & $=$  & $\neq$ & $\neq$ & \cellcolor[gray]{0.8} & \cellcolor[gray]{0.8} &  \cellcolor[gray]{0.8} & \cellcolor[gray]{0.8}  \\
JADE  & $=$   & $=$ & $\neq$ & $\neq$  & $=$  & \cellcolor[gray]{0.8} &  \cellcolor[gray]{0.8} & \cellcolor[gray]{0.8}  \\ 
CoBiDE  & $=$  & $=$ & $\neq$ & $\neq$  & $=$  &  $=$ &  \cellcolor[gray]{0.8} & \cellcolor[gray]{0.8} \\ 
RS  & $\neq$   & $=$  & $\neq$  & $\neq$  & $=$ & $=$ & $=$  &\cellcolor[gray]{0.8}  \\ \hline
\end{tabular}}
\label{t.statistical_caltech}
\end{center}
\end{table}

It is interesting to point out that memory- (IHS) and evolutionary-based (BSA, JADE, and CoBiDE) techniques obtained the best results for all datasets, outperforming swarm collective approaches (AIWPSO, FA, and CS). Regarding evolutionary techniques, mutation and crossover operators may move solutions far apart from each other (i.e., they favor the exploration), which can be interesting in the context of DBM/DBN hyperparameter fine-tuning. Usually, the hyperparameters we are optimizing (i.e., learning rate, number of hidden units, weight decay and momentum)  do not lead to different reconstruction errors under some small intervals, i.e., the fitness landscape figures some flat zones that can trap optimization techniques. 

\begin{table}[!htb]
\caption{Statistical analysis considering Semeion dataset.}
\begin{center}
\resizebox{\columnwidth}{!}{%
\begin{tabular}{l| c c c c c c c c }
\cline{2-9}
                          & IHS    & AIWPSO    & CS   & FA   & BSA    & JADE    & CoBiDE     & RS     \\ \hline
 IHS  &   \cellcolor[gray]{0.8} & \cellcolor[gray]{0.8} &     \cellcolor[gray]{0.8} & \cellcolor[gray]{0.8} & \cellcolor[gray]{0.8} & \cellcolor[gray]{0.8} &  \cellcolor[gray]{0.8}  & \cellcolor[gray]{0.8}  \\ 
 AIWPSO  &  $\neq$  & \cellcolor[gray]{0.8}  &     \cellcolor[gray]{0.8} & \cellcolor[gray]{0.8} & \cellcolor[gray]{0.8} & \cellcolor[gray]{0.8} & \cellcolor[gray]{0.8} &\cellcolor[gray]{0.8}  \\ 
CS &  $\neq$    & $\neq$  &   \cellcolor[gray]{0.8} & \cellcolor[gray]{0.8} & \cellcolor[gray]{0.8} & \cellcolor[gray]{0.8} &  \cellcolor[gray]{0.8} & \cellcolor[gray]{0.8} \\
FA  & $\neq$   &  $\neq$ & $=$  &    \cellcolor[gray]{0.8}  & \cellcolor[gray]{0.8} & \cellcolor[gray]{0.8} &  \cellcolor[gray]{0.8} &\cellcolor[gray]{0.8}  \\
BSA  & $\neq$  &  $=$  &  $\neq$  &  $\neq$ & \cellcolor[gray]{0.8} & \cellcolor[gray]{0.8} &  \cellcolor[gray]{0.8} & \cellcolor[gray]{0.8} \\
JADE  &  $=$  &  $=$ & $\neq$ & $\neq$  &  $=$ & \cellcolor[gray]{0.8} &  \cellcolor[gray]{0.8}  &\cellcolor[gray]{0.8}  \\ 
CoBiDE  & $\neq$   & $=$ & $\neq$ &  $\neq$ &  $=$ & $=$  &  \cellcolor[gray]{0.8} & \cellcolor[gray]{0.8}  \\ 
RS  & $\neq$   & $=$  & $\neq$ &  $\neq$ & $=$  & $\neq$ & $\neq$  &  \cellcolor[gray]{0.8}  \\ \hline

\end{tabular}}
\label{t.statistical_semeion}
\end{center}
\end{table}

Regarding the relatively good results obtained using the Random Search, one may question the contribution of employing metaheuristic techniques for DBM hyperparameter optimization. Despite the statistical similarity among optimization techniques, the random search did not obtain the best results for any dataset.

\subsection{Time Analisys}
\label{ss.timeAnalisys}

Tables~\ref{t.time_mnist},~\ref{t.time_Caltech}, and~\ref{t.time_Semeion} present an analysis of the computational load required by the optimization tasks regarding MNINST, CalTech 101 Silhouettes, and Semeion datasets, respectively. The results in bold stand for the fastest aproaches for each model.

\begin{table}[!htb]
\caption{Computational load (in hours) considering MNIST dataset.}
\begin{center}
\resizebox{\columnwidth}{!}{%
\begin{tabular}{l|l|l|l|l|l|l|l|l|l|l|l|l|}
\cline{2-13}
                          & \multicolumn{4}{c|}{1L} & \multicolumn{4}{c|}{2L} & \multicolumn{4}{c|}{3L} \\ \cline{2-13} 
						  & \multicolumn{2}{c|}{DBN} & \multicolumn{2}{c|}{DBM} & \multicolumn{2}{c|}{DBN} & \multicolumn{2}{c|}{DBM}  & \multicolumn{2}{c|}{DBN} & \multicolumn{2}{c|}{DBM}   \\ \cline{2-13} 
                          & CD    & PCD    & CD   & PCD   & CD    & PCD    & CD   & PCD   & CD    & PCD    & CD   & PCD   \\ \hline
\multicolumn{1}{|c|}{ IHS}  &    0.35 & \textbf{0.25} &     \textbf{0.45} & \textbf{0.46} &   0.60 & 0.52 &   0.57 & \textbf{0.55} & 0.54  & 0.56 & \textbf{0.82} & \textbf{0.53} \\ \hline
\multicolumn{1}{|c|}{ AIWPSO}  & 2.21 & 2.28 &  2.64 & 2.41 &      3.39 & 2.68 &   3.89 & 3.62 & 4.31  & 4.73 & 5.67 & 4.28 \\ \hline
\multicolumn{1}{|c|}{ CS} &      \textbf{0.30} & 0.45 &       0.53 & 0.56 & \textbf{0.49} & \textbf{0.45} &   \textbf{0.44} & 0.80 & \textbf{0.47}  & \textbf{0.29} & 0.84 & 0.97 \\ \hline
\multicolumn{1}{|c|}{ FA}  &     0.75 & 1.49 &      1.81 & 1.06 &  1.37 & 1.30 &   1.95 & 2.41 & 2.23  & 2.22 & 2.52 & 1.29 \\ \hline
\multicolumn{1}{|c|}{ BSA}  &    1.28 & 1.31 &     0.98 & 1.21 &   1.12 & 0.71 &   2.67 & 1.61 & 1.48  & 1.43 & 2.65 & 3.74 \\ \hline
\multicolumn{1}{|c|}{ JADE}  &   1.00 & 1.63 &    0.79 & 0.88 &    1.93 & 1.76 &   2.12 & 1.81 & 1.34  & 1.69 & 3.17 & 2.34 \\ \hline
\multicolumn{1}{|c|}{ CoBiDE}  & 1.25 & 1.29 &  1.11 & 1.11 & 1.50 & 1.67 &    2.13 & 2.22 & 2.29  & 1.60 & 2.92 & 2.26 \\ \hline
\end{tabular}}
\label{t.time_mnist}
\end{center}
\end{table}

One can notice that, in general, IHS has been the fastest technique, followed by CS, which is somehow expected due to their updating mechanism. IHS evaluates a single solution each iteration, while CS evaluates a reduced number of solutions, given by the probability parameter $p$.

\begin{table}[!htb]
\caption{Computational load (in hours) considering  CalTech 101 Silhouettes dataset.}
\begin{center}
\resizebox{\columnwidth}{!}{%
\begin{tabular}{l|l|l|l|l|l|l|l|l|l|l|l|l|}
\cline{2-13}
                          & \multicolumn{4}{c|}{1L} & \multicolumn{4}{c|}{2L} & \multicolumn{4}{c|}{3L} \\ \cline{2-13} 
						  & \multicolumn{2}{c|}{DBN} & \multicolumn{2}{c|}{DBM} & \multicolumn{2}{c|}{DBN} & \multicolumn{2}{c|}{DBM}  & \multicolumn{2}{c|}{DBN} & \multicolumn{2}{c|}{DBM}   \\ \cline{2-13} 
                          & CD    & PCD    & CD   & PCD   & CD    & PCD    & CD   & PCD   & CD    & PCD    & CD   & PCD   \\ \hline
\multicolumn{1}{|c|}{ IHS} &     1.64 & 1.47 &     \textbf{1.81} & \textbf{1.62} &  1.28 & \textbf{1.37} &    \textbf{1.84} &     2.26 & \textbf{1.13}  & \textbf{1.06} &     \textbf{1.98} & 1.58 \\ \hline
\multicolumn{1}{|c|}{ AIWPSO}  & 8.87 & 9.44 & 10.54 & 11.50 &    9.41 & 7.79     & 12.30 & 12.34 &  11.17  & 7.95 & 13.50 & 13.82 \\ \hline
\multicolumn{1}{|c|}{ CS} &      \textbf{1.55} & \textbf{1.01} &      1.86 & 1.63 & \textbf{0.93} & 1.76 &     2.45 &    \textbf{2.17} & 1.46  & 1.35 &     2.00 & \textbf{0.80} \\ \hline
\multicolumn{1}{|c|}{ FA}  &     3.38 & 5.27 &     6.03 & 3.00 &  6.25 & 3.27 &    7.26 &     2.62 & 3.58  & 8.08 &     6.55 & 8.83 \\ \hline
\multicolumn{1}{|c|}{ BSA}  &    6.40 & 5.08 &    6.55 & 8.30 &   6.04 & 5.60 &   9.19 &      8.42 & 4.23  & 4.53 &     7.95 & 9.90 \\ \hline
\multicolumn{1}{|c|}{ JADE}  &   8.24 & 4.31 &   9.22 & 7.90 &    7.71 & 4.10 &  11.15 &      7.40 & 8.25  & 4.57 &    9.43 & 8.29 \\ \hline
\multicolumn{1}{|c|}{ CoBiDE}  & 5.64 & 5.28 & 7.48 & 7.02 & 5.64 & 5.36 & 7.52 & 7.61 &     4.47  & 5.38 &      6.63 &  8.70 \\ \hline
\end{tabular}}
\label{t.time_Caltech}
\end{center}
\end{table}

Likewise, one can expect that BSA, JADE, and CoBiDE to behave similarly regarding the computational load, since they are evolutionary-based techniques and the number of new solutions (the ones that employ mutation and crossover operations) to be evaluated depends upon a probability.

\begin{table}[!htb]
\caption{Computational load (in hours) considering Semeion Handwritten Digit dataset.}
\begin{center}
\resizebox{\columnwidth}{!}{%
\begin{tabular}{l|l|l|l|l|l|l|l|l|l|l|l|l|}
\cline{2-13}
                          & \multicolumn{4}{c|}{1L} & \multicolumn{4}{c|}{2L} & \multicolumn{4}{c|}{3L} \\ \cline{2-13} 
						  & \multicolumn{2}{c|}{DBN} & \multicolumn{2}{c|}{DBM} & \multicolumn{2}{c|}{DBN} & \multicolumn{2}{c|}{DBM}  & \multicolumn{2}{c|}{DBN} & \multicolumn{2}{c|}{DBM}   \\ \cline{2-13} 
                          & CD    & PCD    & CD   & PCD   & CD    & PCD    & CD   & PCD   & CD    & PCD    & CD   & PCD   \\ \hline
\multicolumn{1}{|c|}{ IHS}  &    \textbf{0.16} & 0.19 &    \textbf{0.22} & \textbf{0.25} &     \textbf{0.23} & \textbf{0.20} &     \textbf{0.22} & 0.31 & \textbf{0.28}  & 0.28 & \textbf{0.35} & \textbf{0.38} \\ \hline
\multicolumn{1}{|c|}{ AIWPSO}  & 1.14 & 1.00 & 1.49 & 1.44 &        1.61 & 1.41 &     2.04 & 1.98 & 2.15  & 1.80 & 2.51 & 2.45 \\ \hline
\multicolumn{1}{|c|}{ CS} &      0.26 & \textbf{0.18} &      0.31 & 0.26 &   0.26 & 0.24 &     0.31 & \textbf{0.20} & \textbf{0.28}  & \textbf{0.23} & 0.40 & \textbf{0.38} \\ \hline
\multicolumn{1}{|c|}{ FA}  &     0.49 & 0.74 &     0.82 & 0.42 &    0.62 & 0.98 &     0.82 & 0.53 & 0.84  & 1.14 & 0.90 & 0.76 \\ \hline
\multicolumn{1}{|c|}{ BSA}  &    0.68 & 0.65 &    0.57 & 0.88 &     0.54 & 0.44 &     0.57 & 1.16 & 0.83  & 0.92 & 1.30 & 1.51 \\ \hline
\multicolumn{1}{|c|}{ JADE}  &   0.54 & 0.22 &   0.92 & 1.18 &      0.25 & 0.80 &     0.92 & 1.60 & 0.37  & 1.29 & 1.91 & 2.09 \\ \hline
\multicolumn{1}{|c|}{ CoBiDE}  & 0.71 & 0.58 & 0.74 & 0.91 & 0.49 & 0.89 &  0.96 & 1.03 & 0.68  & 1.01 & 1.52 & 1.30 \\ \hline
\end{tabular}}
\label{t.time_Semeion}
\end{center}
\end{table}

One shortcoming of FA and AIWPSO concerns their computational burden since every agent in the swarm generates a new solution to be evaluated at each iteration. In fact, they are expected to present a slower convergence than IHS, which creates a single solution instead (i.e., it evaluates the fitness function only once per iteration). Such behavior makes them much faster than swarm-based techniques, but having a slower convergence as well.

%% file: sections/conclusion.tex
\section{Conclusions}
\label{s.conclusions}

In this work, we dealt with the problem of fine-tuning Deep Boltzmann Machines by means of meta-heuristic-driven optimization techniques to reconstruct binary images. The experimental results over three public datasets showed the validity in using such techniques to optimize DBMs when compared against a random search. Also, we showed DBMs can learn more accurate models than DBNs considering two out of three datasets. Moreover, we provided a detailed analysis of the similarity among each optimization technique using the Wilcoxon signed-rank test, as well the trade-off between the computational load demanded by each metaheuristic and its effectiveness.

Even though all techniques have obtained close results, we observed that evolutionary- and memory-based approaches might be more suitable for DBM/DBN fine-tuning hyperparameters. Since we are coping with hyperparameters that, under small intervals, do not influence the learning step (i.e., the reconstruction error), evolutionary operators and the process of creating new harmonies seem to introduce some sort of perturbation that moves possible solutions far apart from each other. In regard to future works, we aim to validate the proposed approach to reconstruct and also classify gray-scale images.